\documentclass[runningheads]{llncs}
\usepackage{graphicx}

\usepackage{tikz}
\usepackage{comment}
\usepackage{amsmath,amssymb} 
\usepackage{color}
\usepackage[accsupp]{axessibility}  

\usepackage[linesnumbered,noend, ruled]{algorithm2e}
\usepackage{amsmath}
\usepackage{amsfonts}
\usepackage{algpseudocode}
\usepackage{textcomp}
\usepackage{xcolor}
\usepackage{stackengine}
\usepackage{subcaption}
\usepackage{makecell}
\stackMath
\begin{document}
\pagestyle{headings}
\mainmatter
\def\ECCVSubNumber{4872}  

\title{Bayesian Optimization with Clustering and Rollback for CNN Auto Pruning} 

\titlerunning{BO with Clustering and Rollback for CNN Auto Pruning}
\author{Hanwei Fan\inst{1,2}\and
Jiandong Mu\inst{2}\and
Wei Zhang\inst{2} }
\authorrunning{H. Fan et al.}
%
\institute{The Hong Kong University of Science and Technology (Guangzhou)\and
The Hong Kong University of Science and Technology
\email{\{hfanah,jmu\}@connect.ust.hk, wei.zhang@ust.hk}}
\maketitle

\begin{abstract}
Pruning is an effective technique for convolutional neural networks (CNNs) model compression, but it is difficult to find the optimal pruning policy due to the large design space. To improve the usability of pruning, many auto pruning methods have been developed. Recently, Bayesian optimization (BO) has been considered to be a competitive algorithm for auto pruning due to its solid theoretical foundation and high sampling efficiency. However, BO suffers from the curse of dimensionality. The performance of BO deteriorates when pruning deep CNNs, since the dimension of the design spaces increase.
We propose a novel clustering algorithm that reduces the dimension of the design space to speed up the searching process. Subsequently, a rollback algorithm is proposed to recover the high-dimensional design space so that higher pruning accuracy can be obtained. 
We validate our proposed method on ResNet, MobileNetV1, and MobileNetV2 models. Experiments show that the proposed method significantly improves the convergence rate of BO when pruning deep CNNs with no increase in running time. The source code is available at https://github.com/fanhanwei/BOCR.
\keywords{CNN, Pruning, Bayesian Optimization}
\end{abstract}

\section{Introduction}
Convolutional neural networks (CNNs) are becoming popular due to their high performance and universality. There is a growing trend to apply CNNs in different scenarios such as object detection, speech recognition, \emph{etc}. However, the high performance of CNNs is at the expense of their large model size and high computing complexity, which have prevented them from having broader usage. To solve this problem, network pruning \cite{han2015learning} has been proposed to reduce the model size with little accuracy loss. Many works, \emph{e.g.}, \cite{guo2016dynamic,he2017channel,louizos2017learning,wen2016learning,zhuang2018discrimination}, have been proposed to prune CNNs with different granularity. Among these works, channel pruning \cite{he2017channel,wen2016learning}, which reduces the model size and computing complexity by removing the redundant channels on the feature map, is widely adopted due to its high efficiency in hardware implementation. 

As the depth of CNNs rapidly increases, the design space of the pruning policies, which indicates the preservation ratio of each layer of the CNN model, becomes too large to be fully explored by handcrafted efforts. To reduce the manpower overhead introduced by the pruning process while exploring the design space of pruning, reinforcement learning (RL) \cite{he2018amc,yu2021auto} and general probabilistic algorithms \cite{liu2020autocompress,liu2019metapruning,Zhang_2021_ICCV} are utilized to automate the channel pruning process. 
However, the above methods lead to a large time overhead as they need massive data and training trials to converge. To increase the practicality of auto pruning, a better algorithm is expected to search the design space more efficiently. 

Bayesian optimization (BO) \cite{mockus1978application} is an effective method for tuning the hyper-parameters for the black-box function with high sample efficiency. Therefore, BO is considered to be a competitive candidate algorithm for building the automatic pruning agent \cite{chen2018constraint,ma2019bayesian,tung2017fine}.
However, BO suffers from a fatal drawback that the sampling efficiency of BO drops significantly when dealing with high-dimensional problems. Thus, it would be challenging to prune very deep networks with the BO agent. Currently, the BO-based pruning framework is usually used to deal with shallow networks to maintain high efficiency, and an enhanced BO agent is needed to provide better results when dealing with modern CNN models. Although many algorithms \cite{kandasamy2015high,ma2020additive,rana2017high,wang2016bayesian,wang2013bayesian} have been proposed to mitigate the performance degradation of high-dimensional BO, these works only provide general solutions based on theoretical analysis, which might not be practical for specific applications. When applying BO to CNN pruning, specialized methods can be developed based on our prior knowledge about CNN pruning.

In this work, observing that some CNN layers have similar redundancy, we propose to cluster the layers by exploiting the similarity of their intrinsic properties and train the BO agent in a low-dimensional space.
However, the dimensionality reduction risks missing the optimal pruning policy since the low-dimensional BO does not explore the whole design space. To achieve optimum, we propose a rollback algorithm in which we recover the original high-dimensional searching space for the BO agent and perform a fine-grained search with the low-dimensional data as the prior knowledge. In addition, to fully utilize the information collected during the policy searching in a low-dimensional space, we propose an adaptive searching domain scaling scheme to reduce the workload of the BO agent after rollback so that a faster fine-grained search can be achieved. Our proposed methods not only improve the performance of the BO agent significantly but enjoys simple implementation.

In summary, we make the following contributions:
  \begin{enumerate}
    \item We propose to solve the high-dimensional problem with a clustering-based dimension reduction scheme so that the BO agent can prune CNNs efficiently.
    \item A rollback algorithm is used to recover the high-dimensional space so that the optimal pruning policy will not be missed, and the accuracy of the pruned model can be further improved. 
    \item Experiments show that our methods explore the design space with a considerable improvement in accuracy than naive BO with no increase in running time. When pruning ResNet56, our method delivers a 2.2\% higher accuracy. For challenging tasks like pruning MobileNetV1 and MobileNetV2 on ImageNet, our method achieves a 2.0\% and 1.9\% higher accuracy, respectively.
  \end{enumerate}

\section{Related Work}
\label{background}
\subsection{CNN Pruning}
CNN model pruning has become a heated topic as CNNs are widely used in resource-constrained devices. A significant number of research works have been proposed to prune CNNs by removing the unimportant weights \cite{han2015learning,molchanov2016pruning,hu2016network,srinivas2015data}. Recently, many research works focused on layer-wise channel pruning as it can achieve competitive performance while being hardware-friendly\cite{he2017channel,wen2016learning}.
However, determining the optimal preservation ratio of each layer for the input models is challenging even for experts. In \cite{he2018amc}, the authors employed a deep deterministic policy gradient (DDPG) agent \cite{lillicrap2015continuous}, which is one of the most popular RL algorithms for continuous action spaces, to automatically generate the optimal pruning policy for channel pruning so that human efforts can be released from the tedious handcrafted work. In \cite{wang2019haq}, this RL-based auto pruning scheme was further extended to model quantization for the first time. In the recent work \cite{yu2021auto}, the graph encoder and decoder were applied to generate the states for the RL, which further improved the learning outcome. Other works utilized probabilistic algorithms \emph{e.g.}, simulated annealing \cite{liu2020autocompress}, evolutionary \cite{liu2019metapruning} and MCMC \cite{Zhang_2021_ICCV}, to sample the pruning policy.  Both RL and probabilistic algorithms need massive iterations to converge, causing the pruning process to be time-consuming. Therefore, it is of great interest to develop a better auto pruning agent to improve the convergence speed.

\subsection{BO-based Auto Pruning and High-dimensional BO}
BO is an optimization framework that employs a continuously updated probabilistic model to predict the performance and variation of the design space so that sampling efficiency can be maximized. Nowadays, BO is widely used to tune hyperparameters, and it has become a natural thought to apply BO to CNN pruning. A few works \cite{chen2018constraint,ma2019bayesian,tung2017fine} on automatically pruning CNN based on a BO agent are introduced here.
In \cite{tung2017fine}, the author proposed a fine-pruning method, which applied a BO agent to automatically adapt the layer-wise pruning parameters over time as the network changes. The work in \cite{chen2018constraint} further improved this framework by setting constraints on BO and designing a cooling scheme to prune the CNN model to a user-specified preservation ratio gradually. However, these works dismissed the curse of high dimensionality of BO as they only conducted experiments on shallow networks such as AlexNet \cite{krizhevsky2012imagenet}. 
In \cite{ma2019bayesian}, the authors successfully applied BO to prune deeper networks with an efficient acquisition function and a fast quality measure of the sampled network. However, the high-dimensional problem of the BO agent was not fundamentally solved. 
As an open problem, high-dimensional BO has attracted the attention of many researchers. \cite{rana2017high} found that the length scale, a hyperparameter of Gaussian kernel that controls the smoothness, significantly impacts the performance of high-dimensional BO. Therefore, the author proposed an algorithm to tune the length scale in each iteration. However, the algorithm was only validated on squared exponential (SE) kernel, which is known to be unrealistic for modeling many physical processes \cite{rasmussen2003gaussian}. Besides, there are two mainstream solutions for high-dimensional BO. One of them is additive-GP \cite{kandasamy2015high,ma2020additive,rolland2018high,wang2017batched}, which assumes an additive structure of the target function, making it not applicable to layer-wise pruning. The other is the random embedding approach \cite{letham2020re,qian2016derivative,wang2016bayesian,wang2013bayesian}, which maps the high-dimensional problem to an efficient subspace with the assumption that the unimportant dimensions can be replaced by the combinations of the important dimensions. Although the recent research \cite{Wang_2021_CVPR} showed that the important layers might exist, obtaining the proper embedding is very difficult for random embedding methods. 

Considering the above research gaps, we propose a novel clustering algorithm to reduce the dimensionality based on a moderate assumption that similar layers can share the same preservation ratios. Then, a rollback algorithm is followed to further boost the accuracy.

\section{Methodology}
\label{methods}
In this section, we introduce our methodology by forming the auto pruning task as a BO process. The frequently used variables are shown in Table \ref{tab:vars}.
\begin{table}[b]
    \centering
    \resizebox{0.7\columnwidth}{!}{
    \begin{tabular}{|c|c|}
      \hline 
      variable & meaning \\
      \hline
      \hline
      $i$ & index of layer \\
      \hline
      $n$ & number of input channels for layer \\
      \hline
      $k$ & kernel size \\
      \hline
      $c, c'$ & number of output channels before/after pruning \\
      \hline
      $t$ & the $t^{th}$ iteration of BO process \\
      \hline
      $\Theta, \Theta'$ & network model before/after pruning \\
      \hline
      $p_i$ & preservation ratio for layer $i$  \\
      \hline
      $p_{target}$ & target preservation ratio \\
      \hline
      $N$ & number of layers of the network \\
      \hline
      $\textbf{P}$ & pruning policy for the model \\
      \hline
    \end{tabular}}
    \caption{Variables in this work.}
    \label{tab:vars}
\end{table}

\subsection{Channel Pruning with BO}
In this work, we mainly focus on channel pruning as it can achieve a good trade-off between model size and accuracy while being hardware-friendly. We adopt the magnitude-based channel pruning scheme along with the weight reconstruction method proposed in \cite{he2017channel} to prune the neural network. However, our proposed framework also can be applied to other pruning schemes. 
In channel pruning, a weights tensor with the shape of $n \times c \times k \times k$ is pruned into $n \times c' \times k \times k$, so the preservation ratio $p$ is $c'/c$. Then, the problem becomes determining the optimal $p_i$ for layer $i$ to maximize the accuracy of the pruned network while satisfying the constraints, which can be formulated as the following optimization problem: 
\begin{align}
\begin{split}
    \max_{\textbf{P}} \ & f(\Theta') \\
    s.t.\ & p_f \le p_{target} \\
        \ & \Theta' = Pruning(\Theta, \textbf{P}) \\
        \ & p_f = Flops(\Theta') / Flops(\Theta),
\end{split}
\end{align}
where the $Pruning$ and $Flops$ functions are well-defined and can be implemented explicitly in the program. $f$ is the target function, which is usually a black-box function. In our case, it is the accuracy of the pruned network and can be measured by conducting inferences on the images in the validation set.

This problem is hard to solve since there is no explicit form of $f$ due to the complex relationship between channel pruning policy and the corresponding accuracy. An alternative method is to build a fast model to approximate the black box function $f$ by iteratively interacting with the channel pruner, and the optimal pruning policy $\mathbf{P}^*$ can be achieved by finding the maximum of the model. BO, which models the black-box function with a continually updated probabilistic model, \emph{e.g.} Gaussian Process (GP) model \cite{mockus1991bayesian}, becomes promising for optimizing expensive black-box functions due to its high sample efficiency. 

During the BO process, we note the sampled policy in the $t$th iteration as $\textbf{P}_t \in \mathbb{R}^N$, where $N$ is the depth of the network. Similarly, the pruning policies for time $1$ to $t$ can be noted as $\textbf{P}_{1:t}$. As we have mentioned before, we assume that the pruning samples can be modeled as a GP model, and therefore we have
\begin{equation}
f \left(\Theta, \mathbf{P}_{1:t} \right) \sim \mathcal{N} \left( \mathbf{m} ( \mathbf{P}_{1:t} ), \mathbf{K} \left( \mathbf{P}_{1:t}, \mathbf{P}_{1:t} \right) \right),
\end{equation}
where $\mathbf{m}$ is the mean function, and $\mathbf{K}( \mathbf{P}_{1:t}, \mathbf{P}_{1:t} )$ is the variance matrix. In the following discussion, we denote $f(\Theta, \mathbf{P}_{1:t})$ by $f(\mathbf{P}_{1:t})$ for simplicity. Then, the joint distribution of the previous samples together with the next sample can be represented by
\begin{align}
    \left[ \stackanchor{ f ( \mathbf{P}_{1:t} )}{ f(\mathbf{P}_{t+1} ) } \right] \sim \mathcal{N} \left( 
    \stackanchor{\mathbf{m}(\mathbf{P}_{1:t})}{\mathbf{m}(\mathbf{P}_{t+1})}, 
    \left[ \stackanchor{\mathbf{K}(\mathbf{P}_{1:t},\mathbf{P}_{1:t}), \  \mathbf{k}(\mathbf{P}_{1:t},\mathbf{P}_{t+1}) }{ \mathbf{k}(\mathbf{P}_{t+1},\mathbf{P}_{1:t}), \  k(\mathbf{P}_{t+1},\mathbf{P}_{t+1}) } \right] \right)
\end{align}
and the probabilistic prediction of the next sample can be obtained by
\begin{equation}
\label{eq:var_mean}
\begin{split}
    f(\mathbf{P}_{t+1} ) &\sim \mathcal{N} \left( \mu(\mathbf{P}_{t+1}), \sigma(\mathbf{P}_{t+1}) \right) \\
    \mu(\mathbf{P}_{t+1})  &=\mathbf{k}(\mathbf{P}_{t+1}, \mathbf{P}_{1:t})  \mathbf{K}(\mathbf{P}_{1:t},\mathbf{P}_{1:t})^{-1} f(\mathbf{P}_{1:t}) \\
    \sigma(\mathbf{P_{t+1}}) &= k(\mathbf{P}_{t+1}, \mathbf{P}_{t+1}) \\
     &- \mathbf{k}(\mathbf{P}_{t+1}, \mathbf{P}_{1:t})\mathbf{K}(\mathbf{P}_{1:t}, \mathbf{P}_{1:t}) \mathbf{k}(\mathbf{P}_{1:t}, \mathbf{P}_{t+1})
\end{split}  
\end{equation}
This means that the mean for the unexplored pruning policy $\mathbf{P}_{t+1}$ can be predicted via the history samples, and the corresponding variance of the prediction can also be obtained.

To enhance the sampling efficiency, a cheap surrogate function, which is called the acquisition function, is built to recommend the next sample point with the highest potential to maximize the objective function. 
Expected Improvement (EI) \cite{brochu2010tutorial,mockus1994application}, which is defined by $\mathbb{E}[\max (f(\mathbf{P}) - f(\mathbf{P}^+), 0)]$, aims to find the sampling point that has the highest expected improvement over the current optimal policy $\mathbf{P}^+$, and has become one of the most popular acquisition functions over the past years. In this work, we utilize EI to sample the pruning policies.
\begin{figure}[t]
\centering
\includegraphics[width=0.85\columnwidth]{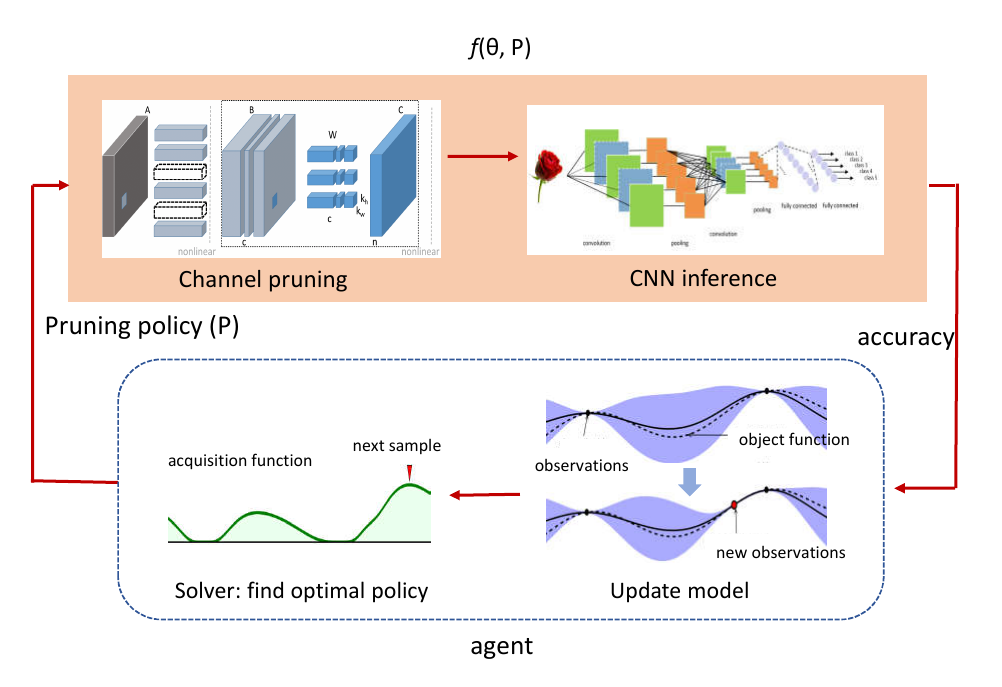}
\caption{Solving the auto pruning problem by BO.}\label{fig:solver_flow}
\end{figure}
The framework for the BO-based auto pruning scenario thus can be illustrated by Fig. \ref{fig:solver_flow}. We first evaluate the randomly generated policies as the initial samples. The initial policies are sampled according to the Sobol sequence \cite{joe2003remark,joe2008constructing} to make the BO process more stable as Sobol is an evenly distributed quasi-random low dependency sequence and has an overwhelming advantage in providing stable and evenly distributed samples.
Then, we build the GP model to estimate the policies' mean and variance at unobserved locations according to equation \ref{eq:var_mean}. Next, the EI-based acquisition function is computed to indicate the potential benefits of each unexplored policy. Finally, the recommended policy, which shows the highest potential of obtaining a better-pruned network, will be given by solving the maximum value of the acquisition functions and will serve as the next sample to update the GP model. By iterating this process, the recommended policy can keep improving until convergence.
Based on the theoretical research of \cite{bull2011convergence}, the simple regret of BO, which defines as $f(\mathbf{P}^*) - f(\mathbf{P}_t)$, is upper bounded by $O(t^{-1/N})$, showing that the convergence rate of BO will significantly decrease as the CNN models get deeper. Therefore, we propose a layer clustering algorithm to solve this problem.

\begin{figure}[t]
\centering
    \begin{subfigure}[b]{0.49\columnwidth}
    \centering
    \includegraphics[width=\columnwidth]{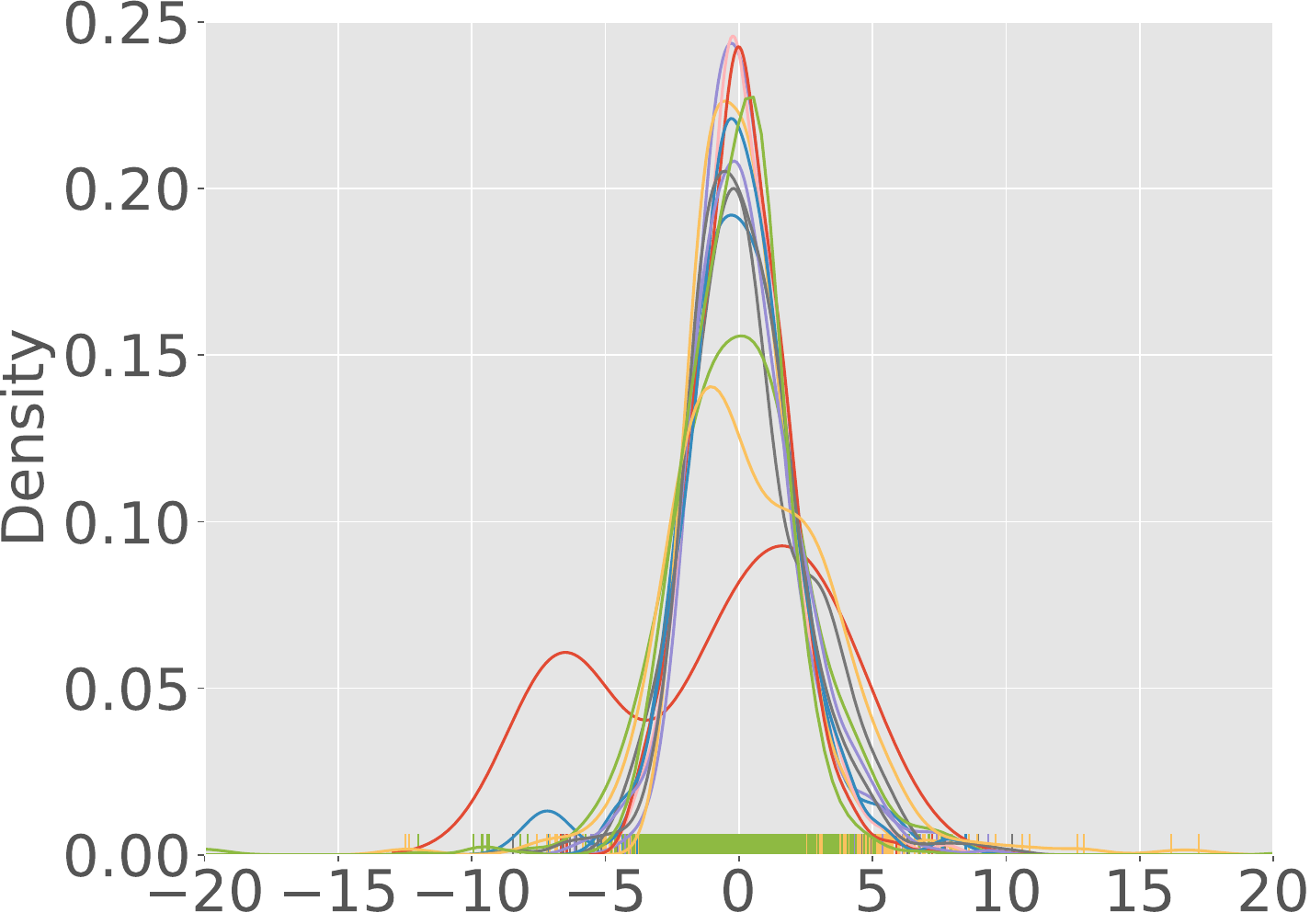}
    \caption{MobileNetV1}
    \label{fig:mobilenetv1D}
    \end{subfigure}
    \hfill
    \begin{subfigure}[b]{0.467\columnwidth}
    \centering
    \includegraphics[width=\columnwidth]{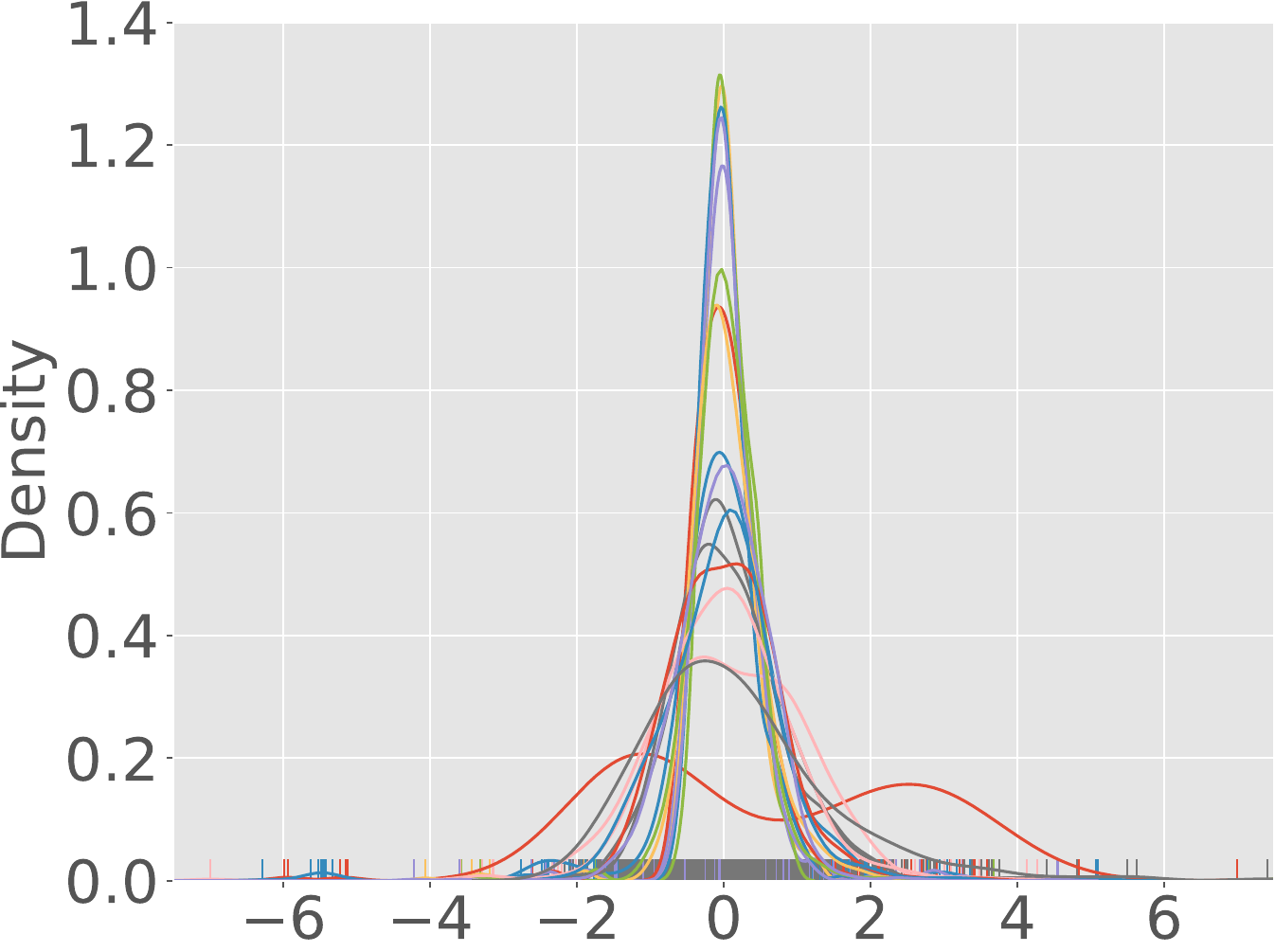}
    \caption{MobileNetV2}
    \label{fig:mobilenetv2D}
    \end{subfigure}
    \caption{The distributions of the channel importance of the layers.}
    \label{fig:mobilenetVD}
\end{figure}
\subsection{Layer Clustering}
To reduce the dimensionality of BO-based auto pruning, we consider clustering the layers and sharing the same preservation ratio within a cluster.
As we frequently observed that several layers have similar preservation ratios, we assume that sharing the same preservation ratio among similar layers will not affect the pruning result. Aiming to exploit the similarity between layers, we propose three different measures and experimentally compare their effectiveness.

Inspired by \cite{he2018amc} which used the layer structure parameters to form the environment states of the RL agent, we define a structural vector and measure the similarity based on the Euclidean distance between the vectors. Our proposed structural vector includes basic parameters \emph{e.g.,} the number of input channels ($n$), number of output channels ($c$), and kernel size ($k$). In addition, we take the number of parameters and Flops into account as they show the computation complexity of a layer. We also include the size and the dimensional change of the output feature map to compare the features extracted by the layers.

Another idea is to compare the distributions of the weights inside the layers. Since we adopt the importance-based pruning scheme that removes the less significant part of the weights, layers with similar distributions are more likely to have close preservation ratios. Note that, although we choose magnitude as the importance in this work, our method can also apply to another kind of importance \emph{e.g.}, gradient-based \cite{hayou2020robust}, Nisp \cite{yu2018nisp}, \emph{etc}. We utilize Gaussian kernel-density estimation (GKde) to fit the Gaussian distribution of each layer's channel importance. The distributions of the layers usually have different biases and do not overlap. Considering whether a channel is redundant depends on its position in the layer it belongs to, layers with different biases can have similar percentages of redundancy. Thus we relocate the distributions by subtracting their medians to make different distributions comparable. Fig. \ref{fig:mobilenetVD} shows the distributions of different layers in MobilenetV1 and MobilenetV2, and it is obvious that some layers have similar channel importance distributions. Next, we compare two different methods to measure the similarity between distributions. One of them is Jensen–Shannon divergence (JSD), known as the symmetric version of KL divergence, which focuses on the information gap. The other is the Euclidean distance which focuses on shape. 
To find the best approach to measure the similarity of the layers, a set of experiments is conducted and will be shown in the experiment section. 

Taking the similarity measures as the distance between the layers, the hierarchical agglomerative clustering (HAC) \cite{ward1963hierarchical} can be adopted to cluster the layers as it has no limitation on the measure of distance while other methods may require the observations of the data. To minimize the total within-cluster variance, we update the distance between clusters with the Ward linkage method, whose details can be found in \cite{mullner2011modern}. Another characteristic of HAC is that the existing clusters will not split when we decrease the number of clusters $C$, and results of different $C$ are stored after one single run. This characteristic is useful to the rollback algorithm, which will be introduced in the next subsection.

After dividing the layers into $C$ clusters, we only need to train a $C$-dimensional GP model instead of the high-dimensional model, improving the upper bound of the simple regret to $O(n^{-1/C})$. We assign the same preservation ratios for each cluster so that the $C$-dimensional vector generated by the low-dimensional model can be extended to the $N$-dimensional pruning policy $\mathbf{P}$.

\subsection{Rollback for Higher Accuracy}
With the layer clustering algorithm, we can efficiently obtain high-quality pruning policies. However, the possibility exists that we miss the optimal pruning policy since the low-dimensional BO does not explore the whole design space. Therefore, we propose to recover the original dimensionality after the low-dimensional model converges so that the whole design space can be reached and the optimal will not be missed. It is difficult for BO to find the peaks of the acquisition function in high-dimensional space\cite{rana2017high}, leading to random-like searching. However, with the data collected in the low-dimensional space, BO can easily locate the peaks and exploit better results when returning to the high-dimensional space as sufficient prior knowledge about the peaks is provided. To achieve this, we propose a rollback algorithm that rebuilds the GP model to search the whole design space with plenty of prior knowledge, which is presented in Alg. \ref{alg:rollback}. 
\subsubsection{Direct Rollback}
A simple way to rollback is to build a $D$-dimensional model directly. To achieve this, we record the pruning policy $\mathbf{P}$ every iteration. Recall that $\mathbf{P}$ is extended from the low-dimensional sample of BO based on the clustering, thus we can recover the dimensionality of BO by rebuilding the GP model with $\mathbf{P_{1:t}}$. Since BO updates by fitting a new GP model every iteration, rebuilding the GP model causes no extra time overhead. 
Our experiments show that our rollback algorithm can further improve the results by discovering better pruning policies that the clustering-based BO can not reach.
\subsubsection{Gradual Rollback}
A more sophisticated way is to gradually rollback to reduce the learning gap between the low and high-dimensional space. To be detailed, a cluster number $C^*$, which satisfies $C < C^* < N$, can be chosen as the bridge stage so that we can first rollback to $C^*$-dimensional space and then rollback to the original dimensionality. As we mentioned in the last subsection, a nice feature of HAC is the consistency of results among different cluster numbers. Therefore, there is no mismatch between the corresponding dimensions after rollback. In addition, as all results of HAC are saved after one single run, we can obtain the clustering result of $C^*$ clusters without extra computation. To build the $C^*$-dimensional GP model, we only need to remove the dimensions that remain in clusters from $\mathbf{P_{1:t}}$. The effectiveness of gradual rollback is shown in the experiment section, and a detailed analysis of different choices of $C^*$ is shown in the supplementary material.

\begin{algorithm}[bt]
{
\SetAlgoLined
\SetKwInput{KwInput}{Input}
\KwInput{$C$-dimensional GP model $M_C$, Original dimensionality $D$,\\ Max iteration $T$}
\KwResult{D-dimensional GP model}
 converge counter = 0\;
 \While{$t$ $<$ $T/2$}{
    Obtain $\mathbf{P}_{t}$ from $M_C$ and Record $\mathbf{P}_{t}$\;
    \If{$f(\mathbf{P}_{t}) > f(\mathbf{P}^+)$}{
        Push $\mathbf{P}_{t}$ to the Best Queue\;
        converge counter = 0\;
    }
    \If{{\rm converge counter} $\ge 20$}{
        break\;
    }
    update $M_C$\;
    converge counter ++\;
 }
 Obtain the $D$-dimensional clustering result\;
 Reform $\mathbf{P}_{1:t}$ based on the $D$-dimensional clustering\;
 Perform adaptive searching domain scaling\;
 Generate the D-dimensional GP model\;
}
\caption{Rollback the Clustered BO}
\label{alg:rollback}
\end{algorithm}
\subsubsection{Adaptive Searching Domain Scaling}
To boost the efficiency of BO in high-dimensional space, we propose an adaptive searching domain scaling scheme, where we shrink the searching domain according to the history data collected during the low-dimensional BO process. 
We store the pruning policies with the highest accuracy in the Best Queue, which has ten entities, as we observe that ten entities can decrease the searching domain to a reasonable extent.
Then, the searching domain of the high-dimensional BO process can be formulated as

\begin{equation}
    \mathcal{D} = \left[ \min_{i,j} \mathbf{P}^{*i}_{j}, \max_{i,j} \mathbf{P}^{*i}_{j} \right]^C,
\end{equation}
where $i$ ranges from $1$ to $10$, indicating the index of the high-performance samples in the queue. $j$ ranges from $1$ to $C$, indicating the index of the clusters.

\section{Experiments}
\label{experiments}
In our experiments, we use GpyOpt \cite{GPyOpt} as the naive BO agent and implement the proposed methods based on it. We adopt Matern5/2 as the GP kernel as recommended in \cite{rasmussen2003gaussian}. The hyperparameters of the kernel are decided by maximum likelihood estimation.
RL agent in \cite{he2018amc} is chosen as a baseline. To make a fair comparison, we adopt the same channel pruning scheme for all the methods. 
The accuracy is estimated based on a random subset of the training set, whose sizes are 5000 and 3000 for Cifar10 and ImageNet, respectively.
We conduct our experiments on several representative CNN model architectures, including ResNet56 trained on Cifar10 \cite{he2016deep}, MobileNetV1  \cite{howard2017mobilenets} and MobileNetV2 \cite{sandler2018mobilenetv2} both trained on ImageNet. 
We run each experiment for 200 epochs and report the mean ($m$) and the standard deviation ($\sigma$) of 10 different seeds.
Note that our research focuses on the optimization process of auto pruning. Therefore, we perform a detailed analysis of the convergence process and all the experiment results are obtained before finetuning.

\subsection{Analysis of Similarity Measures}
To analyze the three proposed similarity measures, we perform layer clustering based on each of them and compare the searching results. Table \ref{tsim} shows the cluster number $C$ and the mean of top-1 accuracy along with the variance. As shown in Table \ref{tsim}, the structure-based measure provides the best result for ResNet56, while Euclidean distance of distributions performs the best for the other two models.

We believe that significant architectural differences between the models lead to this result. ResNet56 has a plain architecture, as it is composed of repeated layers of only three different structures. Therefore, when the dimensionality reduces to three, the structure-based measure can achieve high accuracy while others suffer a considerable loss. Similarly, the structure-based measure achieves high accuracy for MobileNetV1 as it has five layers that share the same structure. However, the other layers of MobileNetV1 have different structures, which limits the effectiveness of the structure-based measure. For MobileNetV2, the distribution-based measure outperforms the structure-based measure by a large margin, showing its superiority in dealing with complex CNN models.

Except for ResNet56, which is unsuitable for distribution-based measures, Euclidean distance provides higher accuracy and lower variance than JSD. We explain this result in terms of the meaning of Euclidean distance and JSD. Euclidean distance focuses on the relative position of the weights, which has a significant impact on the importance-based pruning method. JSD measures the difference in information, which is not closely related to the importance-based pruning. Thus, Euclidean distance better measures the similarity among layers.

To summarize, the structure-based measure is suitable for simple models with repeated layers while the Euclidean distance of distributions is effective for complex models. Therefore, we adopt the structure-based measure for ResNet56 and adopt the Euclidean distance of distributions for MobileNetV1 and MobileNetV2 in the following experiments.
\begin{table}[t]
    \centering
    \resizebox{\columnwidth}{!}{
    \begin{tabular}{p{3.125cm}<{\centering}p{3.125cm}<{\centering}p{3.125cm}<{\centering}p{3.125cm}<{\centering}}
      \hline 
         & ResNet56 & MobileNetV1 & MobileNetV2 \\
      \hline
      $C$ & 3 & 6 & 6 \\
      \hline
      Structure & \textbf{91.68 (0.43)} & 48.46 (1.39) & 50.31 (0.71) \\
      \hline
      JS & 90.78 (0.27) & 47.99 (1.98) & 52.49 (1.27) \\
      \hline
      Euclidean & 88.95 (0.55) & \textbf{48.50 (1.20)} & \textbf{52.57 (0.98)} \\
      \hline
    \end{tabular}}
    \caption{Comparison of Similarity Measures.}
    \label{tsim}
\end{table}

\begin{table}[t]
    \centering
    \resizebox{\columnwidth}{!}{
    \begin{tabular}{p{2.5cm}<{\centering}p{2.5cm}<{\centering}p{2.5cm}<{\centering}p{2.5cm}<{\centering}p{2.5cm}<{\centering}}
      \hline 
      method & top-1 $m$ & top-1 $\sigma$ & top-5 $m$ & top-5 $\sigma$ \\
      \hline
      RL & 87.69 & 3.39 & 99.42 & 0.33 \\
      \hline
      Naive BO & 90.29 & 1.63 & 99.72 &  0.11 \\
      \hline
      Layer Clustering & 91.68 & 0.43 & 99.77 & 0.06 \\
      \hline
      Rollback & \textbf{92.53} & 0.56 & \textbf{99.86} & 0.05 \\
      \hline
    \end{tabular}}
    \caption{Performance for ResNet56.}
    \label{tab:Resnet56}
\end{table}

\subsection{Experiments on ResNet56}\label{sec:Resnet56}
ResNet56 is a representative model architecture trained on Cifar10 and it has a considerably large depth, leading to very high dimensions for layer-wise pruning tasks. Although the first layers of the residual branches are not prunable because the input feature maps are shared with the shortcut branches, there remain 28 layers to be pruned, which will cause the curse of dimensionality problem for the naive BO agent. While with layer clustering, there are only three parameters left for the BO agent to optimize.

In Table \ref{tab:Resnet56}, we list the mean value $m$ and the standard deviation $\sigma$ of both top-1 and top-5 accuracy achieved by different methods when pruning 50\% Flops for ResNet56. Our proposed layer clustering method improves BO's performance with 1.4\% higher top-1 accuracy. The rollback scheme further improves the accuracy by 0.9\%. Note that in \cite{he2018amc}, the author shows that AMC can achieve a 90.2\% top-1 accuracy in 400 epochs. Our proposed rollback method can achieve 93.14 \% within 200 epochs, which is significantly better than \cite{he2018amc}.

\begin{figure}[t]
\centering
    \begin{subfigure}[b]{0.32\columnwidth}
    \centering
    \includegraphics[width=\columnwidth]{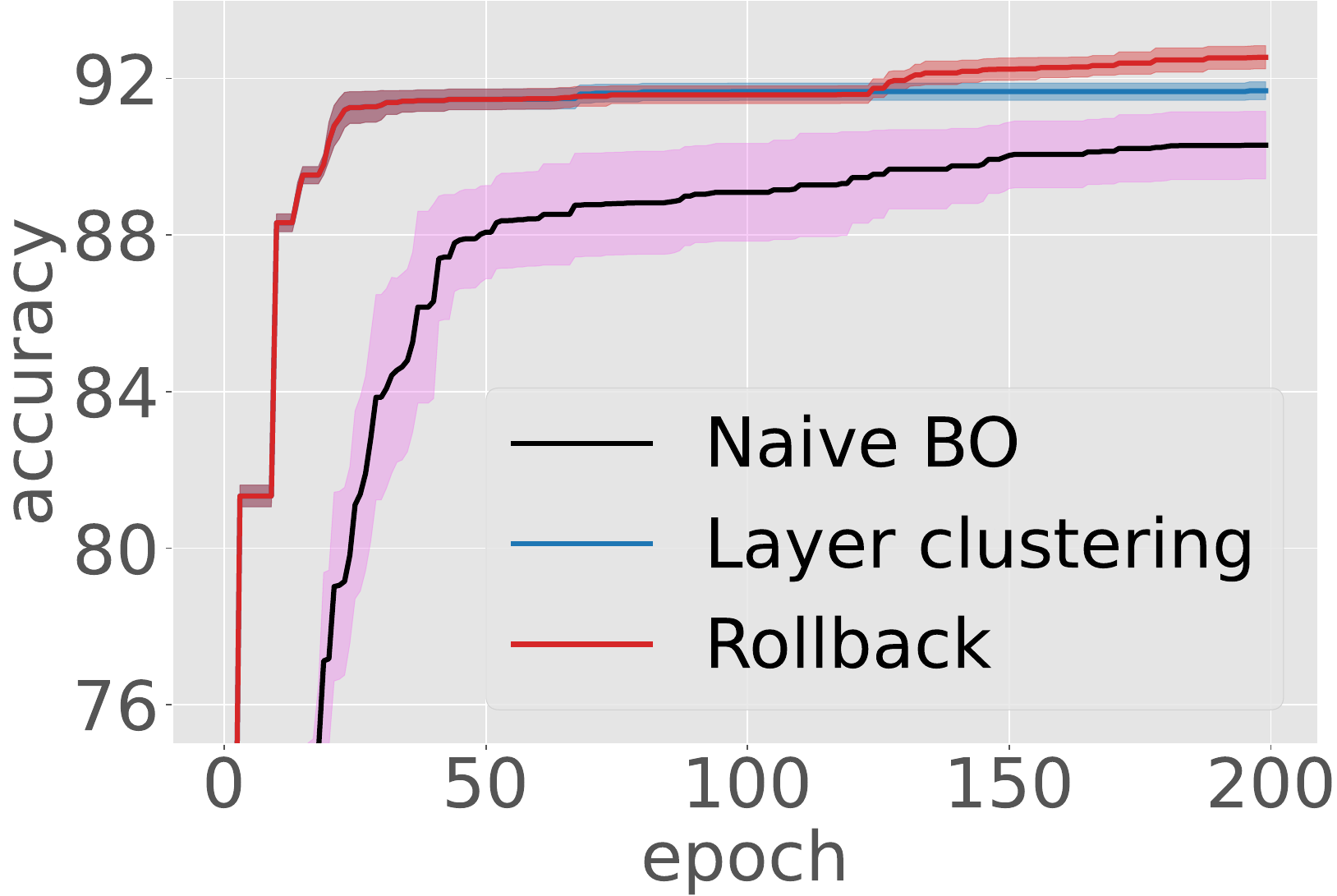}
    \caption{ResNet56}
    \label{fig:res56}
    \end{subfigure}
    \hfill
    \begin{subfigure}[b]{0.32\columnwidth}
    \centering
    \includegraphics[width=\columnwidth]{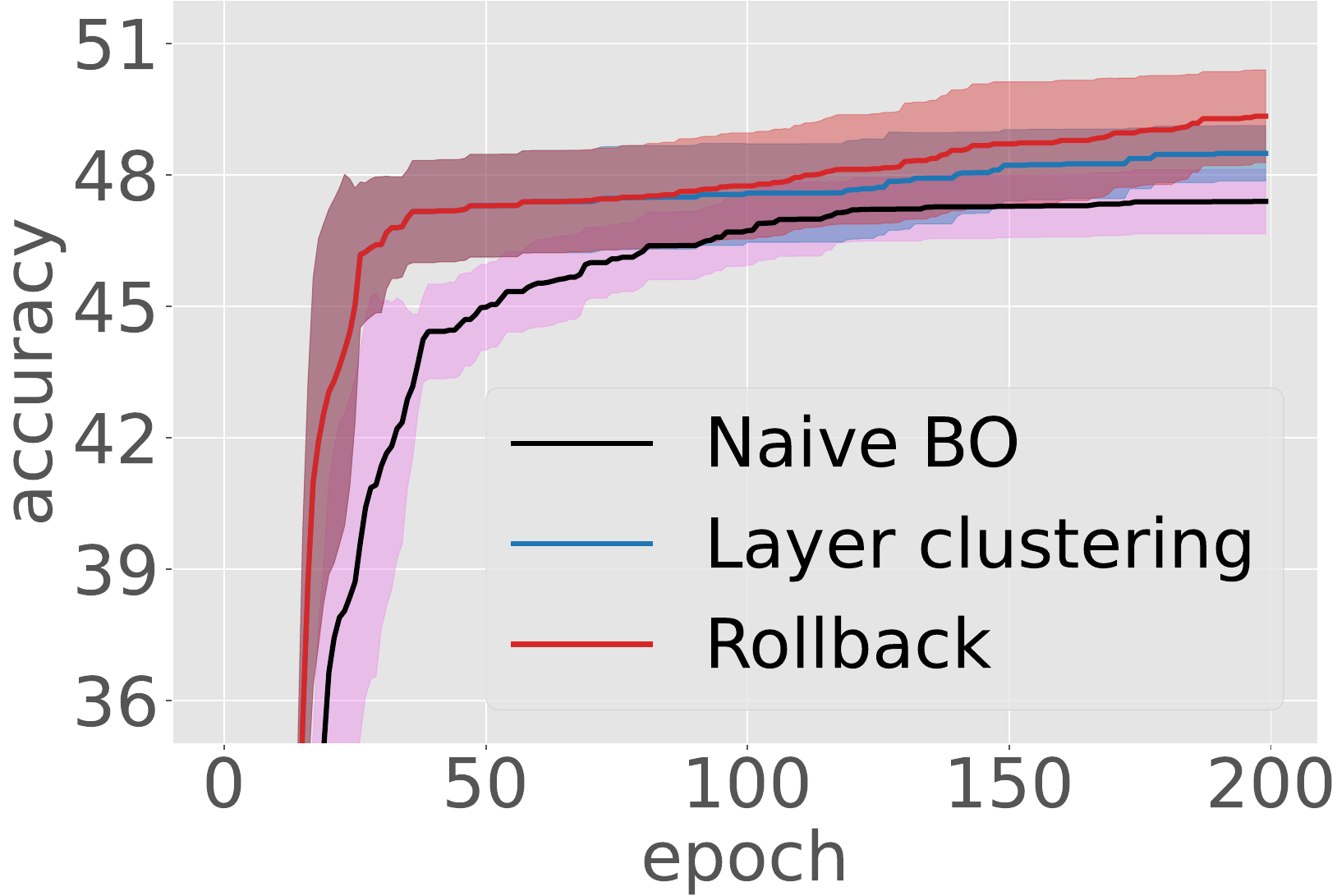}
    \caption{MobileNetV1}
    \label{fig:mobilenetv1}
    \end{subfigure}
    \hfill
    \begin{subfigure}[b]{0.32\columnwidth}
    \centering    
    \includegraphics[width=\columnwidth]{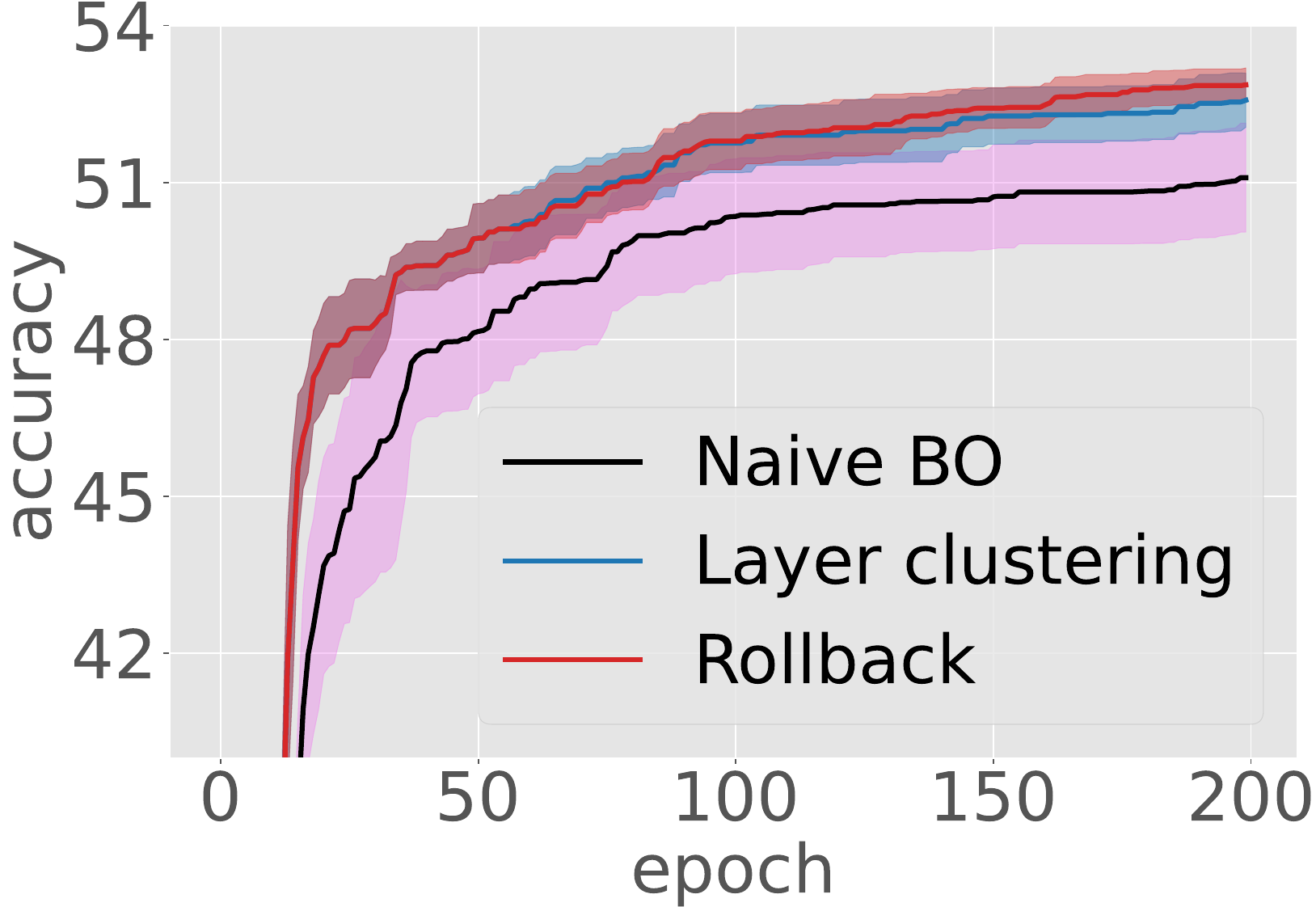}
    \caption{MobileNetV2}
    \label{fig:mobilenetv2}
    \end{subfigure}
    \caption{Comparison of BO-based methods.}
    \label{allexperiments}
\end{figure}

Table \ref{tab:Resnet56} also shows that the BO agent outperforms RL in efficiency by a large margin. It can be observed that the BO-based searching scheme is more stable than the RL-based counterpart and converges much faster, as the $\sigma$ of the RL agent is much higher than our proposed layer clustering and the rollback algorithm. 
In Fig. \ref{fig:res56}, we show the effectiveness of the proposed methods in detail. The solid lines in the figure refer to the means, and the shaded areas refer to the corresponding $\sigma$. The layer clustering method can significantly boost the convergence of the BO agent. After its convergence, the rollback scheme turns the design space back into a high-dimensional space and can further improve the accuracy of the pruned network.

Note that all methods take around 800 seconds to finish 200 epochs in our device, which indicates the time spent for each trial in BO and RL is close and the time overhead of rollback is ignorable. Thus, our method is also much more efficient from the perspective of wall clock time. 

\subsection{Experiments on MobileNetV1}\label{sec:Mobilenetv1}
MobileNetV1 is a popular single-branch network trained on ImageNet and it is known to be challenging to prune due to its compact design. As the layers of MobileNet consist of pairs of depth-wise convolution and point-wise convolution layers, we only consider the point-wise convolution layers when searching for the pruning policy, and the corresponding channels in the depth-wise layer will be removed accordingly. We also note that the first layer is not prunable as its channel should be in line with the input images, and similarly for the final linear. As a result, there are 13 parameters for BO to optimize. We prune 50\% Flops and we divide the layers into 6 clusters.

In Table \ref{tab:MobilenetV1}, we show the accuracy and corresponding $\sigma$ for the proposed methods on MobileNetV1. Similar to the result of ResNet56, BO-based methods achieve significantly better results than the RL agent. Our rollback-based BO can achieve the best top-1 accuracy of 52.7\%, while the best top-1 accuracy of the RL counterparts is 50.2\%.
Additionally, the layer clustering-based BO outperforms the original BO agent by 1.1\% and 0.9\% in top-1 and top-5 accuracy. The rollback scheme further improves the accuracy by 0.9\% and 0.5\% in top-1 and top-5 accuracy, respectively. Note that it is normal that the $\sigma$ gets higher after rollback as the simple regret increases in high-dimensional space.

In Fig. \ref{fig:mobilenetv1}, we compare the three BO-based methods in detail. The layer clustering method speeds up the convergence of the BO agent significantly, based on which rollback scheme further improves the accuracy.
\begin{table}[t]
    \centering
    \resizebox{\columnwidth}{!}{
    \begin{tabular}{p{2.5cm}<{\centering}p{2.5cm}<{\centering}p{2.5cm}<{\centering}p{2.5cm}<{\centering}p{2.5cm}<{\centering}}
      \hline 
      method & top-1 $m$ & top-1 $\sigma$ & top-5 $m$ & top-5 $\sigma$ \\
      \hline
      RL & 45.61 & 1.95 & 71.88 & 1.87 \\
      \hline
      Naive BO & 47.39 & 1.39 & 73.11 & 1.47 \\
      \hline
      Layer Clustering & 48.49 & 1.20 & 74.03 & 0.87 \\
      \hline
      Rollback & \textbf{49.34} & 2.01 & \textbf{74.52} & 1.66 \\
      \hline
    \end{tabular}}
    \caption{Performance for MobileNetV1.}
    \label{tab:MobilenetV1}
\end{table}

\begin{table}[t]
    \centering
    \resizebox{\columnwidth}{!}{
    \begin{tabular}{p{2.5cm}<{\centering}p{2.5cm}<{\centering}p{2.5cm}<{\centering}p{2.5cm}<{\centering}p{2.5cm}<{\centering}}
      \hline 
      method & top-1 $m$ & top-1 $\sigma$ & top-5 $m$ & top-5 $\sigma$ \\
      \hline
      RL & 43.15 & 5.45 & 69.84 & 5.01 \\
      \hline
      Naive BO & 51.09 & 1.97 & 77.13 & 1.33 \\
      \hline
      Layer Clustering & 52.57 & 0.98 & 78.45 & 1.12 \\
      \hline
      Rollback & \textbf{52.86} & 0.61 & \textbf{78.57} & 0.70 \\
      \hline
    \end{tabular}}
    \caption{Performance for MobileNetV2.}
    \label{tab:MobilenetV2}
\end{table}

\subsection{Experiments on MobileNetV2} 
We also validate our method on the modern efficient network MobileNetV2, which is trained on ImageNet. As an improved version of MobileNetV1, MobileNetV2 is even more compact than MobileNetV1, making it challenging to prune. It adopts an inverted residual structure while keeping the depth-wise and point-wise design, leading to complex architecture. We use the same experimental setting as in Sec. \ref{sec:Resnet56} and  \ref{sec:Mobilenetv1} for the residual structure and the depth \& point-wise structure, and there are 18 parameters for the BO agent to optimize. We divide the layers into 6 clusters and preserve 60\% Flops. 

As shown in Fig. \ref{fig:mobilenetv2}, the layer clustering algorithm successfully boosts the convergence of BO, which is consistent with previous experiments. 
Table \ref{tab:MobilenetV2} shows that our layer clustering algorithm can raise the top-1 accuracy over the naive BO agent by 1.5\% while achieving a much lower $\sigma$. Moreover, our method outperforms the RL-based counterpart by a large margin as RL can not converge within the same iterations for this challenging task.

However, the rollback method only improves the top-1 accuracy of layer clustering by 0.29\% which is 3$\times$ lower than the previous experiments. The reason is that the original dimensionality of MobileNetV2 is higher than MobileNetV1, resulting in a learning gap after performing the direct rollback. This motivates us to develop the gradual rollback algorithm. Note that since the image size of Cifar10 is much smaller than ImageNet, the design space of ResNet56 is also smaller. Therefore, the learning gap is alleviated for ResNet56.

\subsection{Gradual Rollback for Higher Dimensionality}
In this section, we demonstrate the effectiveness of the gradual rollback scheme with experiments on MobileNetV2.
Both the structure-based measure and Euclidean distance of distributions are included for the layer clustering phase. To be fair, the bridge stage $C^*$ is 15 for both experiments. 
Table \ref{gradual} shows the {top-1} accuracy achieved by the layer clustering algorithm and compares the improvement provided by rollback and gradual rollback. 
Results show that gradual rollback successfully mitigates the learning gap and outperforms direct rollback.
Additionally, when an inferior measure is chosen for the layer clustering, gradual rollback can make up for it by significantly increasing the accuracy.

\begin{table}[t]
    \centering
    \resizebox{0.85\columnwidth}{!}{
    \begin{tabular}{p{2.5cm}<{\centering}p{2.5cm}<{\centering}p{2.5cm}<{\centering}p{2.5cm}<{\centering}}
      \hline 
      models & Layer clustering & Direct Rollback & Gradual rollback \\
      \hline
      Structure & 51.06 & +0.41 & +1.60 \\
      \hline
      Euclidean & 52.57 & +0.29 & +0.44 \\
      \hline
    \end{tabular}}
    \caption{Comparison between Direct and Gradual Rollback.}
    \label{gradual}
\end{table}

\subsection{Choice of the Cluster Number}
The choice of the cluster number $C$ is a common problem for clustering algorithms. 
In this work, if $C$ is large, the sampling efficiency of BO will decrease. When $C$ is small, good pruning policies are more likely to be excluded from the searching space. For models with plain architecture like ResNet56, the number of different layers they contain is a good choice for $C$. However, there is no obvious choice of $C$ for complex models. To analyze the effect of $C$, we set $C$ to different values and test them on MobileNetV1 and MobileNetV2.
As shown in Table \ref{table:cluster number}, choices around 6 lead to the best results. Note that the architecture of MobileNetV1 and MobileNetV2 are quite different, showing the generality of this result. Therefore, cluster numbers around 6 are reasonable choices. 

\begin{table}[t]
    \centering
    \resizebox{\columnwidth}{!}{
    \begin{tabular}{p{2cm}<{\centering}p{1.5cm}<{\centering}p{1.5cm}<{\centering}p{1.8cm}<{\centering}p{1.8cm}<{\centering}p{1.5cm}<{\centering}p{1.5cm}<{\centering}}
      \hline 
       top1 & 4 & 5 & 6 & 7 & 8 & 9 \\
       \hline
       MobiletNetV1 & 47.65 & 47.12 & \textbf{47.7 (0.65)} & 47.7 (2.19) & 45.82 & 46.27 \\
      \hline
       MobiletNetV2 & 49.48 & 50.6 & \textbf{52.11} & 50.93 & 50.95 & 50.43 \\
      \hline
    \end{tabular}}
    \caption{Comparison of different cluster numbers. In parentheses is the $\sigma$.}
    \label{table:cluster number}
\end{table}

\section{Conclusion}
\label{conclusion}
We have analyzed the similarity between the CNN layers and proposed a novel layer clustering algorithm to boost the sampling efficiency of BO for CNN pruning. To further improve the accuracy of the output pruning policy, we developed a rollback algorithm to recover high-dimensional design space and perform a fine-grained search with the data collected in the low-dimensional space as the prior knowledge. Our experiments have validated the effectiveness of the proposed algorithms. In the future, we intend to extend our work to other layer-wise pruning methods that measure the importance of channels with different metrics. We are also interested in the theoretical proof for the rollback algorithm. 

\section{Acknowledgments}
We would like to thank the anonymous reviewers for their valuable comments. We also thank the Turing AI Computing Cloud (TACC) \cite{TACC} and HKUST iSING Lab for providing us computation resources on their platform. This research was supported in part by Hong Kong Research Grants Council General Research Fund (Grant No. 16215319).
\clearpage
%
%
\bibliographystyle{splncs04}
\bibliography{ref}
\end{document}